\documentclass{article}

     \usepackage[preprint]{neurips_2019}

\usepackage[utf8]{inputenc} % allow utf-8 input
\usepackage[T1]{fontenc}    % use 8-bit T1 fonts
\usepackage{hyperref}       % hyperlinks
\usepackage{url}            % simple URL typesetting
\usepackage{booktabs}       % professional-quality tables
\usepackage{amsfonts}       % blackboard math symbols
\usepackage{nicefrac}       % compact symbols for 1/2, etc.
\usepackage{microtype}      % microtypography

\usepackage{hyperref}
\usepackage{amsmath}

\usepackage[pdftex]{graphicx}

\title{Radial Prediction Layers}

\author{%
  Christian Herta and Benjamin Voigt \\
  Faculty 4 / Center for Bio-Medical image and Information processing (CBMI)\\
  University of Applied Science HTW\\
  Berlin, 12459 \\
  \texttt{christian.herta@htw-berlin.de}, \texttt{benjamin.voigt@htw-berlin.de} \\
}

\begin{document}

\maketitle

\begin{abstract}

For a broad variety of critical applications, it is essential to know how confident a classification prediction is. In this paper, we discuss the drawbacks of softmax to calculate class probabilities and to handle uncertainty in Bayesian neural networks. We introduce a new kind of prediction layer called \textit{radial prediction layer} (RPL) to overcome these issues. In contrast to the softmax classification, RPL is based on the open-world assumption.
Therefore, the class prediction probabilities are much more meaningful to assess the uncertainty concerning the novelty of the input. We show that neural networks with RPLs can be learned in the same way as neural networks using softmax. On a 2D toy data set (spiral data), we demonstrate the fundamental principles and advantages. On the real-world ImageNet data set, we show that the open-world properties are beneficially fulfilled. Additionally, we show that RPLs are less sensible to adversarial attacks on the MNIST data set. Due to its features, we expect RPL to be beneficial in a broad variety of applications, especially in critical environments, such as medicine or autonomous driving.

\end{abstract}

\section{Introduction}\label{sec:introduction}

Even though deep neural networks are successfully applied to a variety of classification problems~\citep{LeCun2015}, it was shown that the predictions are often fragile and tiny changes in the input could lead to an erroneous classification~\citep{Szegedy2013, Su2019}.

Especially in critical environments (e.g., medical applications and autonomous driving), it is necessary to know how confident respectively, how uncertain a (classification) prediction is~\citep{Gal2016Uncertainty, Begoli19}.
Predictive uncertainty can be categorized into two principle types~\citep{NIPS2017_7141}. \textit{Aleatoric} uncertainty captures noise inherent in the observations. \textit{Epistemic} uncertainty accounts for uncertainty in the model, which can be explained away given enough data. An approach to handle the epistemic uncertainty is the use of Bayesian methods~\citep{Hinton:1993, Blundell15, Gal16}. Here, we focus on a specific type of epistemic uncertainty which arises if an input at test time is quite different from all of the training examples.

In this paper, we investigate softmax as classification layer in traditional neural networks and Bayesian neural networks and discuss the drawbacks of softmax. We argue that if softmax is used the uncertainty in the weights (epistemic uncertainty) cannot appropriately capture the uncertainty for examples quite different from training data (novel input ${\bf x}$) if the mean field approximation is used, section~\ref{Bayesian neural networks}.

To overcome this issue, we implemented a simple alternative that we call \textit{radial prediction layer} (RPL). In contrast to softmax, RPL relies on the open-world assumption~\cite[chapter~5]{2018Chen}, i.e. the sum of the prediction probabilities for all classes can be significantly smaller than one, e.g., because a test example belongs to a new class which was not in the training data set. This paper describes the fundamental principles and advantages of RPLs on a 2D toy data set (spiral data) for visualization purposes. It demonstrates that the new prediction layer is as flexible as softmax, i.e., it can be used with any type of neural network structure (recurrent, convolutional, etc.) to compute classification probabilities. Neural networks with RPL for the prediction can be trained by a frequentist or by a Bayesian approach with the same optimization criteria as softmax networks.

We further investigated RPLs on the real-world ILSVRC data set (ImageNet~\citep{Deng09}) using standard deep neural networks architectures. This paper shows that in such networks, the open-world properties are beneficially fulfilled and similar results can be achieved, as with architectures using softmax. Additionally, it describes that RPLs are less sensible to adversarial attacks on the MNIST data set.

An implementation of RPL based on PyTorch and NumPy is available at:

\url{https://gitlab.com/peroyose/radial_prediction_layers}

\section{Related Work}
Our approach focuses on the uncertainty which results from test data dependent on the novelty of the input.
Different techniques exist to detect novel data, see, e.g., \cite{Chandola09} for a summary.
\cite{Bishop94} used radial basis functions for density estimation to identify novel examples (input) for neural networks.
However, such methods typically do not scale to high dimensional data which are typical for deep learning applications.

Another approach for handling novelty in classification with neural networks is to introduce prediction layers which rely on the open-world assumption~\citep{Bendale15, Shu2017}.
% OpenMax
\cite{Bendale15} propose an extension to softmax, which the authors call \textit{openmax}. It is based on extreme value theory. To get this open-world extension an additional term in the partition function for a non-class is computed and used. It was also shown that  \textit{openmax} is more robust against adversarial examples~\citep{DBLP:conf/bmvc/RozsaGB17}. We show a similar result with our RPLs, see section~\ref{sec:adversarial}.
% DOC: Deep Open Classification of Text Documents
The approach by Shu et al. is called \textit{deep open classification} (DOC). DOC builds a 1-vs-rest predication layer based on \textit{sigmoids} rather than softmax and applies a Gaussian fitting to improve the decision boundaries~\citep{Shu2017}.

\section{Softmax and Uncertainty}

Usually, classification probabilities are computed by softmax. softmax relies on the closed world assumption, i.e. each data example should be classified to one of the predefined classes $y \in \{1, \dots, K\}$ ($K$ is the number of  classes). For the $D$-dimensional input ${\bf x} \in \mathbb R^{D}$ the probability that ${\bf x}$ belongs to the class $j$ is denoted by $p(y=j \mid {\bf x})$.
%~\footnote{We defined the input as $D$-dimensional vector. However, the input can also be an image or a sequence.}. 
With the unnormalized outputs $o_j$ of the neural network $p(y=j \mid {\bf x})$ is computed by

\begin{equation}\label{eq:softmax}
  p(y=j \mid {\bf x}) = \frac{\exp(o_j)}{\sum_{i}\exp(o_i)}
\end{equation}

\begin{figure}[hbt!]
  %centering
  \includegraphics[width=0.45\linewidth]{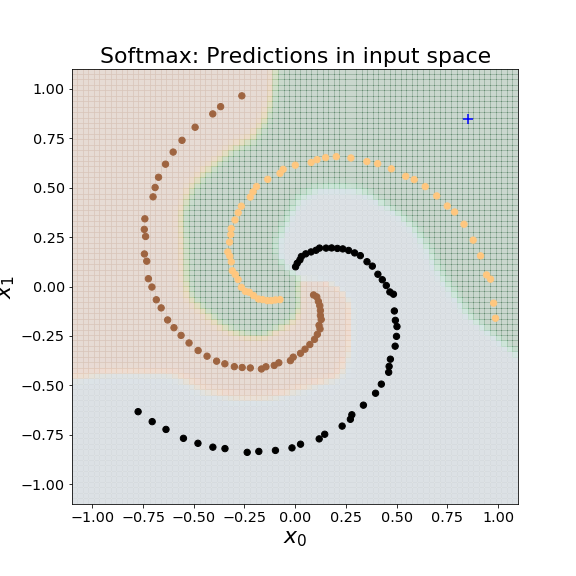}
   \includegraphics[width=0.45\linewidth]{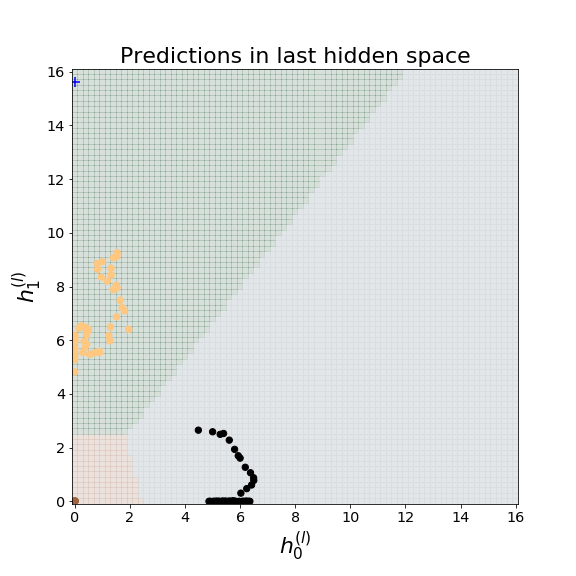}
%  \caption{On the left side the spiral data set is shown. There are three classes (different colors) corresponding to three spirals.
%    The blue cross represents a test example which is ``far away'' from the training data. For such test examples the uncertainty
%    of the prediction should be very high.
%    On the right side the last hidden layer before the softmax layer is shown. In this layer the network maps the data such that the
%    data is linear separable. Due to the use of ReLU non linearities only the positive quadrant is accessible (and shown here).}
  \caption{On the left side the spiral data set and a typical prediction by a neural network with a softmax layer is shown. The blue cross represents a test example which is ``far away'' from the training data. For such test examples the uncertainty of the prediction should be very high.
However, a high probability for  the yellow class is predicted by softmax.  
   On the right side the last hidden layer before the softmax layer is shown. The network maps the data into this layer such that the data is linear separable. Due to the use of ReLU non-linearities only the positive quadrant is accessible (and shown here).}
  %On the right hand side a typical decision by a neural network with a radial prediction layer is shown. White background color indicates that all classes have low predicted probability.}
  \label{fig:spiral_data_decisions}
\end{figure}

By definition of softmax, the predicted probabilities for all classes sum up to one. As we discuss later, the closed world assumption is problematic if at test time an input ${\bf x}$ is quite different from all the training examples. We call such inputs "novel" or ``far away'' from the training data, for an example see the blue data point in figure~\ref{fig:spiral_data_decisions}. Formally, a data point ${\bf x}$ is "novel" if the probability for sampling such a data point ${\bf x}$ as a training example is approximate zero, i.e., $p_{train}({\bf x})\approx 0$.

%An analogon from physics can motivate softmax. The probability of a system to be in a state $s_j$ is proportional to the negative exponentiated energy $\exp(-E(s_j))$ (Boltzmann distribution). Normalization with the partition function $Z = \sum_{i} \exp(-E(s_i))$ results in probability $p(s_j) = \exp(-E(s_j))/Z$ ~\citep{reif2009fundamentals}. Based on this interpretation the neural network learns a mapping from the input ${\bf x}$ to a (negative) log-energy space ${\bf o}$, i.e. ${\bf x}\rightarrow {\bf o}$. Softmax transforms the negative log-energies in class prediction probabilities.

%\begin{figure}
%  %centering
%  %\includegraphics[width=0.45\linewidth]{./pics/spiral-data.png}
%  \includegraphics[width=0.45\linewidth]{./pics/spiral_data_last_hidden_softmax.png}
%  \caption{On the left side the spiral data set is shown. There are three classes (different colors) corresponding to three spirals.
%    The blue cross represents a test example which is ``far away'' from the training data. For such test examples the uncertainty
%    of the prediction should be very high.
%    On the right side the last hidden layer before the softmax layer is shown. In this layer the network maps the data such that the
%    data is linear separable. Due to the use of ReLU non linearities only the positive quadrant is accessible (and shown here).}
%  \label{fig:spiral_data}
%\end{figure}

\subsection{Traditional Neural Networks}

We use the spiral data set for illustration purposes. A fully connected feed forward neural network maps the input ${\bf x}$ in each layer to a new representation ${\bf h}^{(i)}$, $i$ being the layer index. With softmax the representation in the last hidden layer ${\bf h}^{(l)}$ has to be linear separable w.r.t. the different classes. So the neural network learns such a mapping ${\bf x} \rightarrow {\bf h}^{(l)}$, see figure~\ref{fig:spiral_data_decisions}(right). That corresponds to complex decision boundaries in the input space, see figure~\ref{fig:spiral_data_decisions} (left). 

\subsection{Bayesian Neural Networks}\label{Bayesian neural networks}

Scalable Bayesian neural networks rely on the mean field approximation and can, e.g., be trained by the backpropagation algorithm minimizing the evidence lower bound (ELBO) in a variational approach~\citep{Blundell15}, see also supplementary material. However, different approximation techniques for learning and prediction for Bayesian networks exist~\citep{Hinton:1993, Blundell15, Gal16}. As well as traditional neural networks, Bayesian neural networks have typically a softmax layer to compute the class probabilities.

In a Bayesian neural network, the mapping from the input to the last hidden layer ${\bf x} \rightarrow {\bf h}^{(l)}$ is not deterministic. ${\bf h}^{(l)}({\bf x})$ is represented by a probability density function $p({\bf h}^{(l)}({\bf x}))$ for a fixed input ${\bf x}$. That is part of the uncertainty of the model (epistemological uncertainty). This uncertainty is mediated by the probability distribution of the weights. Additionally, there is a (probabilistic) mapping from ${\bf h}^{(l)}$ to the logit output ${\bf o}$ which is also part of the model uncertainty. In the mean field approximation these uncertainties are independent. In other words, the probability density factorizes with a term for each unit (and therefore also layer). The probabilistic decision boundaries of the softmax layer are independent of the probabilistic mapping ${\bf x} \rightarrow {\bf h}^{(l)}$.

\begin{figure}[hbt!]
  %centering
  \includegraphics[width=0.45\linewidth]{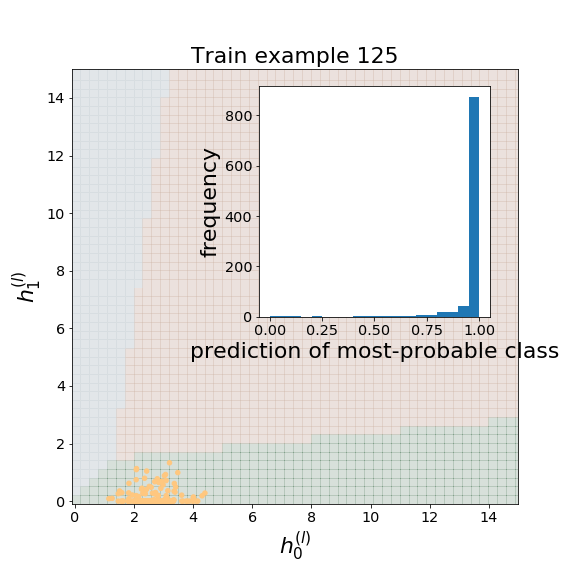}
  \includegraphics[width=0.45\linewidth]{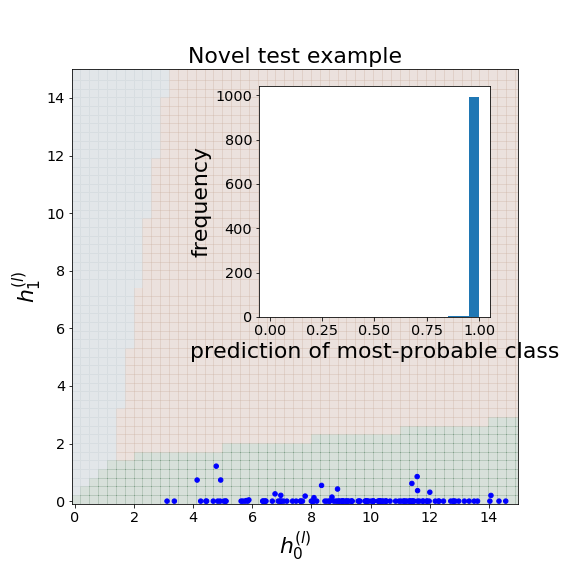}
  \caption{Stochastic mapping from an input data example by different samples of a Bayesian neural network.
  In the inner plot the prediction distribution of the most-probable class for the corresponding data point is shown.
On the left side for a training example. On the right side for the ``far away'' test example (blue cross of figure~\ref{fig:spiral_data_decisions}).
Depending on the stochasticity of the training, e.g. the weight initialization, this can result
in quite high prediction probabilities of the test example.
 }
 \label{fig:last_hidden_bayes}
\end{figure}

In practice, samples of the weights are drawn which corresponds to a sample of the Bayesian network~\footnote{see supplementary material for details.}. If a training example is mapped (${\bf x} \rightarrow {\bf h}^{(l)}$) by different samples of a Bayesian network it must be mapped in the same region. Because of the independence, there cannot be any correlation which adapts the decision boundaries for the specific sample of the Bayesian network. (On the spiral data set,) this results in quite the same decision boundaries for different samples of the Bayesian network (for a figure see supplementary material). The mapping ${\bf x} \rightarrow {\bf h}^{(l)}$ for the training data must be such that the correct class prediction is mainly fulfilled. It is mapped mostly on the correct side of the decision boundaries. So, an input ${\bf x}$ is mapped in nearby regions. The same holds if novel data is mapped in the last hidden layer. If the probabilistic mapping ${\bf x}\rightarrow{\bf h}^{(l)}$ puts the data in a region where high probabilities by softmax logistic regression are assigned then the uncertainty of the prediction seems to be very low. This is not a wanted behaviour, see figure~\ref{fig:last_hidden_bayes}(left)~\footnote{For illustration purpose, the dimensionality of the hidden space was two. However, this result also holds if the dimensionality of the last hidden layer is much higher. We verified this experimentally.}. So, the combination of the softmax model and the mean field approximation can result in strong underestimated (prediction) uncertainties.

\section{Radial Prediction Layers}

A \textit{radial prediction layer} is the last layer of a neural network. As in softmax networks the input ${\bf x}$ is mapped into an output (vector) ${\bf o} = {\bf o}({\bf x})$. In case of RPLs, we call the corresponding vector space the RPL-(vector)-space. % and we denote ${\bf o}^{(rdl)}$ for clarity.
The mapping from the last hidden layer representation ${\bf h}^{(l)}$ to ${\bf o}$ is done by a pure affine transformation without a non-linear activation function. Therefore, the full RPL space is accessible and not e.g., only the positive part (if ReLU would be used). 

In the RPL-space, there is a prototype vector ${\bf p}_j$ for each class $j$. The predicted class probability for an input feature vector ${\bf x}$ is given by $p(y=j \mid {\bf x}) = \exp(- \beta d_j({\bf x}))$, with $d_j({\bf x})$ being the distance of ${\bf o}({\bf x})$ to the prototype vector ${\bf p}_j$. $\beta>0$ is a hyperparameter. As a metric, we use the Euclidean distance, but in principle other distances are possible without restricting the general idea. So with the prototype vector ${\bf p}_{j}$ for class $j$  in the RPL-space and $\mid\mid . \mid\mid_2$ for the notation of the $\ell$2-norm the distance is

\begin{equation}\label{eq:euklidean}
   d_j ({\bf x})= \mid\mid {\bf p}_{j} - {\bf o} ({\bf x}) \mid\mid_2
\end{equation}

If an input is mapped exactly to a prototype then the probability that it belongs to the corresponding class is $\exp(0)=1$. With increased distance the probability goes exponentially towards zero.

We put the prototypes on the axes of the coordinate system of the RPL-space ${\bf p}_j = a {\bf e}_j$. The vectors ${\bf e}_j$ form an orthonormal basis of the coordinate system of the RPL space. %, see figure~\ref{fig:fixed_prototypes} (left).
In neural network terminology, ${\bf e}_j$ is equivalent to a one-hot encoded state representation in the RPL-space. $a$ is the distance of all prototypes to the origin. %, see figure~\ref{fig:fixed_prototypes}. 
The distance $c$ between the prototypes is given by the Pythagorean theorem: $c =\sqrt{2} a$.

The open-world assumption demands that the sum of the predicted probabilities of all classes must be equal or smaller than one, $\sum_k p(y=k \mid {\bf x}; {\bf w}) \leq 1$. If an input is now mapped exactly to a prototype $j$ then the predicted class probabilities for the other classes $k \neq j$ must be zero.
This could be realized by setting the probability $p(y=k \mid {\bf x}; {\bf w})$ to zero if the distance is greater than a threshold $c' \leq c$:

\begin{equation} \label{equ:rpl_prob}
 p(y=k \mid {\bf x}; {\bf w}) =
 \begin{cases}
 	\exp \left(- \beta d_k({\bf x})\right) & \text{if } d_k({\bf x})<c' \\
 	0 & \text{else}
 \end{cases}
\end{equation}

In practice, we just set the hyperparameters $a$ and $\beta$ such that $\exp \left(- \beta c \right) $ has a very small value.

\subsection{Learning}

Without considering regularization, the weights ${\bf w}$ of a neural network are typically learned with a train data set $\mathcal D_{train}=\{\left({\bf x}^{(1)}, y^{(1)}\right), \left({\bf x}^{(2)}, y^{(2)}\right), \dots, \left({\bf x}^{(m)}, y^{(m)}\right)$ by minimizing the negative log-likelihood $- \ell({\bf w}) = - \sum_i \log p(y^{(i)} \mid {\bf x}^{(i)}; {\bf w}) $.

Using the threshold rule~\ref{equ:rpl_prob} for learning would be problematic. If the distance to the target prototype is greater  then $c'$ this would result in a zero gradient. So, during learning we just used  $p(y=k \mid {\bf x}; {\bf w}) = \exp \left(-\beta d_k({\bf x})\right)$. The corresponding (negative log likelihood) loss for RPL of an example $i$ with target class $k$ is

\begin{equation} \label{eq:rpl_loss}
 - \log p(y^{(i)}=k \mid {\bf x}^{(i)}; {\bf w}) = \beta d_k({\bf x}) = \beta  \; \mid\mid  {\bf o}({\bf x}^{(i)})- {\bf p}_k \mid\mid_2
\end{equation}

Minimizing the negative log-likelihood is equivalent to minimizing the distance to the target prototype. The hyperparameter $\beta$ just scales
the log-likelihood and can therefore be neglected in the minimization.

\subsection{Example: Spiral Data}

\begin{figure}[hbt!]
  %centering
  \includegraphics[width=0.9\linewidth]{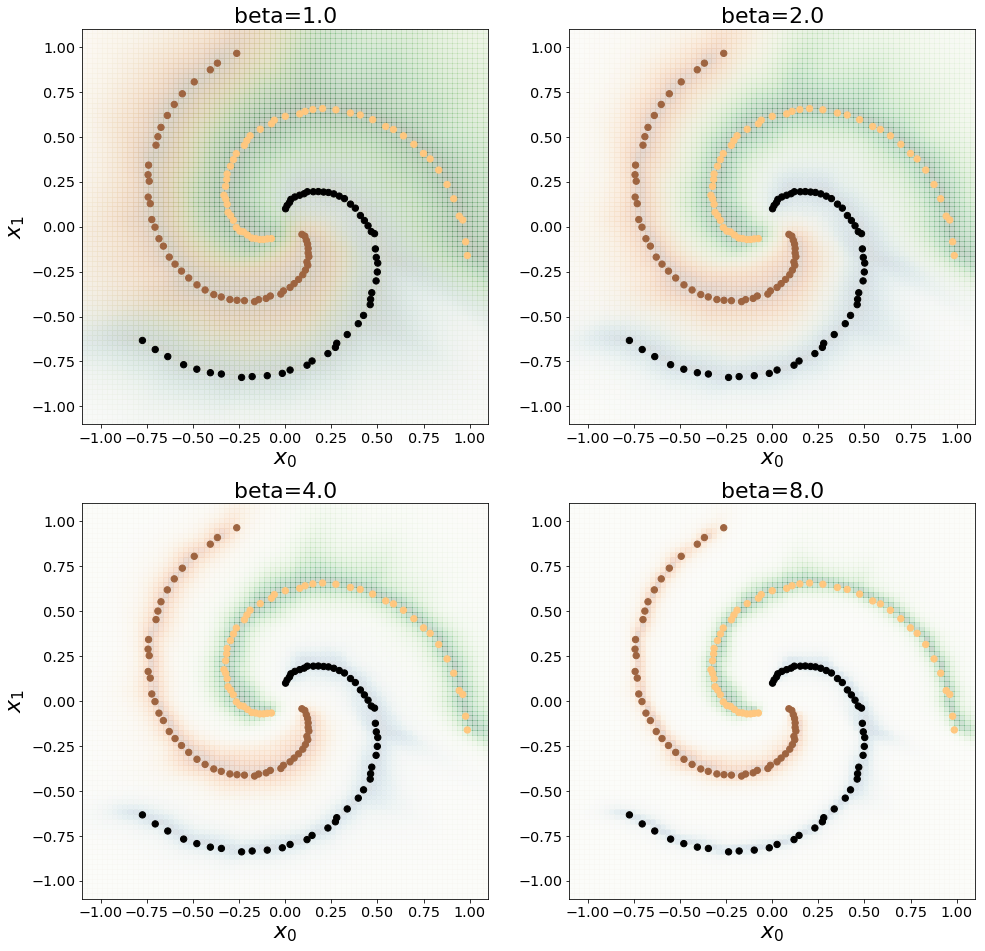}
  \caption{Predictions for different values of the hyperparameter $\beta$ on the spiral data set ($a=1$). With increased value of $\beta$ the predictions with high class probabilities become more and more restricted to the narrow spiral manifolds.}
  \label{fig:spiral_data_rpl_betas}
\end{figure}

In figure~\ref{fig:spiral_data_rpl_betas} (right) typical predictions for the spiral data set are shown with $a=1$. For larger $\beta$-values, only on the spiral manifolds the prediction probabilities for the corresponding classes are substantially greater than zero. In all other regions of the input space the prediction for
all classes is very low. 

\begin{figure}[hbt!]
  %centering
  \includegraphics[width=0.45\linewidth]{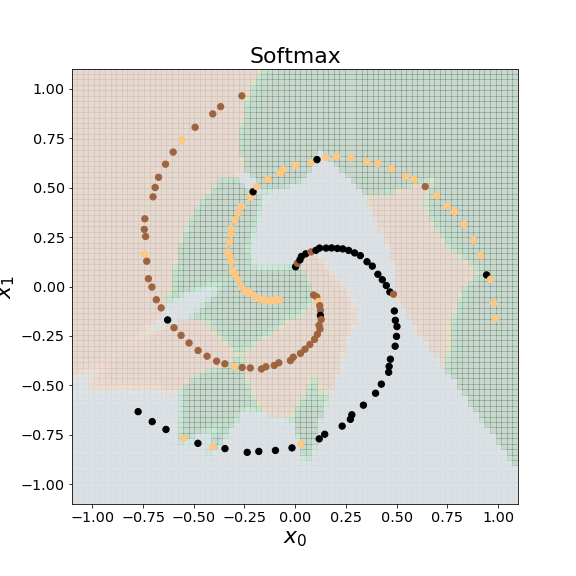}
  \includegraphics[width=0.45\linewidth]{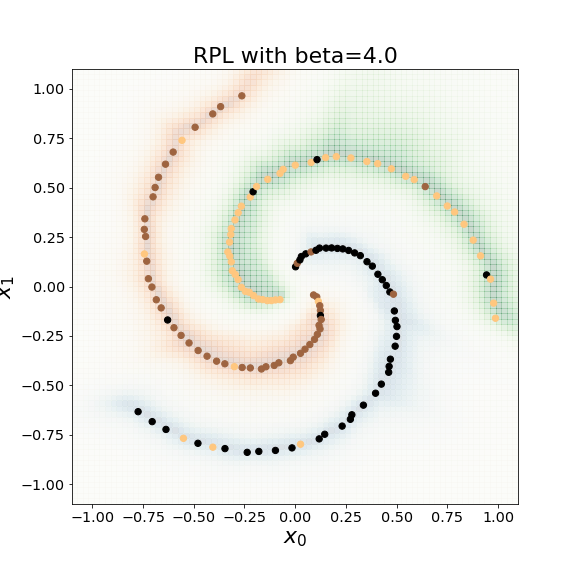}
  \caption{Prediction with noise (10\%), i.e. 10\% of the data points have a random label. High capacity networks with softmax show strong overfitting behaviour (left plot). Surprisingly, high capacity RPL networks seem to be quite robust against overfitting (right plot) with similar predictions as in the no-noise case.}
  \label{fig:spiral_data_noise}
\end{figure}

A result of the spiral data set with noisy labels $t$  is shown in figure~\ref{fig:spiral_data_noise} for softmax and RPL.

\begin{figure}[hbt!]
  %centering
  \includegraphics[width=0.45\linewidth]{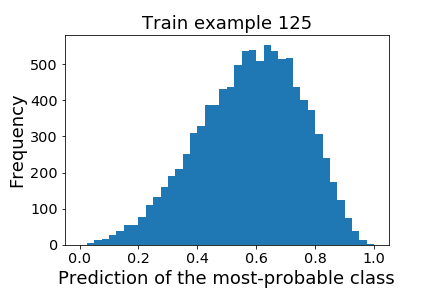}
  \includegraphics[width=0.45\linewidth]{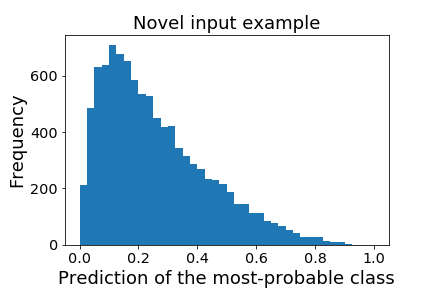}
  \caption{Probability distributions of Bayesian neural networks with RPL analog to figure~\ref{fig:last_hidden_bayes} (inner plots) for the same training (left) and test (right) example. The distributions are much broader and more meaningful as with softmax.}
  \label{fig:spiral_data_bayesian_dist}
\end{figure}

RPL layers can also be used in Bayesian neural networks. We trained such networks with the spiral data set.
In figure~\ref{fig:spiral_data_bayesian_dist} are the prediction probability distributions of a training and a (novel) test example shown.

\subsection{Example: MNIST}

\begin{figure}[hbt!]
  \centering
  \includegraphics[width=0.45\linewidth]{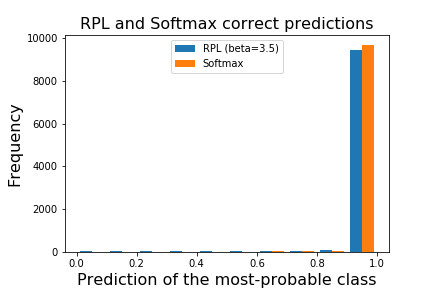}
  \includegraphics[width=0.45\linewidth]{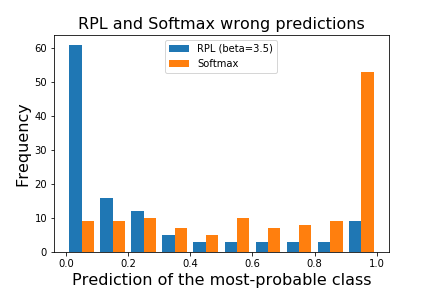}
  \caption{Histograms of the frequencies of correct predictions and wrong predictions (max of the class probabilities) of a neural network with RPL and Softmax as classification layer for the MNIST test data set. In the RPL case, 
for the wrong predictions the probability is mostly quite low.}
  \label{fig:MNIST_class_props}
\end{figure}

We trained both variants without tuning of any hyperparameters (no regularization) on the MNIST train set with a convolutional neural network. Both variants, softmax, and RPL attained a similar accuracy of approximated
 $0.988$. The histograms of the predicted most-probable class for the correct and wrong predictions are shown in figure~\ref{fig:MNIST_class_props}. Sometimes, the predicted class probabilities are used as confidence measures. With softmax, such confidences are even for wrong predictions quite high. In contrast, with RPL wrong predictions have mostly a quite low predicted probability (~\ref{fig:MNIST_class_props}, right).

\subsection{Example: ILSVRC Data}
We explored RPLs on the ILSVRC dataset (ImageNet)~\citep{Deng09} to verify that the approach is applicable to real-world data and deep neural network architectures. We examined several architectures~\citep{krizhevsky2012AlexNet, Simonyan2014VGG, He2016ResNet}, and all could be trained with RPLs. However, this paper focuses on the VGG network~\citep{Simonyan2014VGG}, which we have studied most intently. We trained the network multiple times with one training setup in two variations: \textit{pre-trained} (reinitialize fully connected layers) and \textit{from scratch} (reinitialize all parameters). Additionally, we trained a network on a subset of the ImageNet data, and used the removed classes as novel data in the test phase. A detailed description of network architectures, training setups and hyperparameter is provided in the supplementary material.

\begin{figure}[hbt!]
  \centering
  \includegraphics[width=0.45\linewidth]{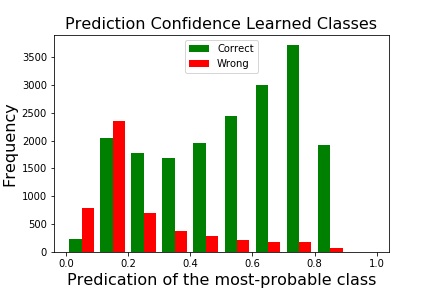}
  \includegraphics[width=0.45\linewidth]{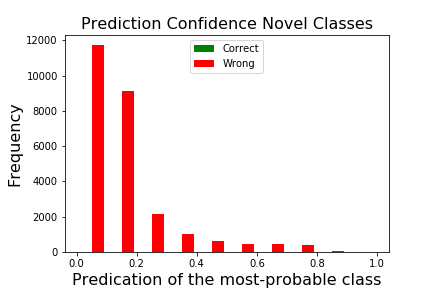}
  \caption{Histogram of the frequencies of correct predictions and wrong predictions of a VGG network with RPL for the validation data (right) and novel data (left) are shown (ImageNet). The network was trained on a subset. Data removed from the training set was used as novel data (novel classes). For the wrong predictions (max of the class probabilities) the probability is mostly quite low for RPL. Note, that by increasing the hyperparameter $\beta$ the distributions could be shifted to the left side.}
  \label{fig:imagenet_class_props_rpl_novel}
\end{figure}

The experiments show that similar accuracies can be achieved with RPLs in deep network architectures compared to softmax. In this statement, we assume an error value(\textit{doubled corrected sample standard deviation}) of X.Y. Table~\ref{tab:accuracy} summarizes the accuracies and compares them to the results achieved by~\citep{Simonyan2014VGG}(Table 3 - \textit{ConvNet performance at a single test scale}) with softmax. Figure~\ref{fig:imagenet_class_props_rpl_novel} presents the results for the network trained on an ImageNet subset. The network was validated on classes used during training and is quite confident in the decisions if we use the predicted class probabilities as confidence measure (right). In contrast, the prediction confidence for novel data is quite low (left) as expected from the open-world property of the RPL.

\begin{table}[hbt!]
  \caption{Achived accuracies for the MNIST and ImageNet datasets. For Imagenet the trained variations VGG (A), (B) and (C) are compared with respect to top-1 and top-5 accuracy. VGG(A) is a pre-trained network with reinitialized fully connected layers using RPL. VGG(B) is a fully reinitialized network using RPL. VGG(C) is the original network reported in~\citep{Simonyan2014VGG}.}
  \label{tab:accuracy}
  \centering
  \begin{tabular}{lccccc}
    \toprule
          & \multicolumn{2}{c}{MNIST}   & VGG (A) & VGG (B) & VGG ~\citep{Simonyan2014VGG}\\
    \cmidrule(r){1-6}
                  & Softmax & RPL       & RPL     & RPL     & Softmax \\
    \midrule
    Acc (top-1)   & 0.9878  & 0.9880    & 76.1    & 78.7     &  74.5\\
    Acc (top-5)   & -       & -         & 91.4    & 90.0     &  92.0\\
    \bottomrule
  \end{tabular}
\end{table}

\section{Adversarial Examples}\label{sec:adversarial}

\begin{figure}[hbt!]
  \centering
  \includegraphics[width=0.45\linewidth]{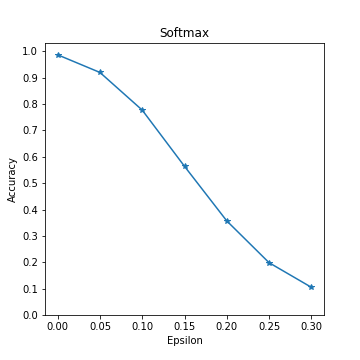}
  \includegraphics[width=0.45\linewidth]{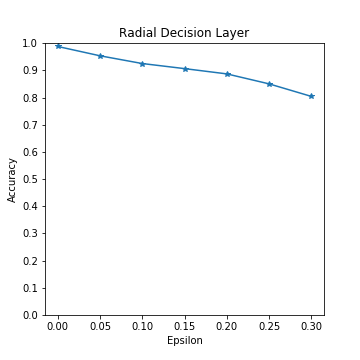}
  \caption{Accuracy of the MNIST test data set for different levels of adversarial perturbation (epsilon).
    On the left for a neural network with a softmax prediction layer and on the right for a neural network with RPL.}
  \label{fig:MNIST_accuracy_vs_epsilon}
\end{figure}

We also investigated how prone RPL is against Fast Gradient Sign Attack (FGSM)~\citep{Goodfellow14} in comparison with softmax on the MNIST dataset. In FGSM the input is additively perturbed in the direction that maximizes the loss. This pushes down the predicted probability of the corresponding class. The strength of the perturbation is controlled by a parameter $\epsilon$ .
With increased $\epsilon$ the accuracy of the test data set drops and more and more test examples are classified incorrectly, see figure~\ref{fig:MNIST_accuracy_vs_epsilon}. From the figure can be seen that RPL is much less prone to FGSM.

\section{Conclusion}

In this paper, we describe the drawbacks of softmax in the context of uncertainty on a novel input ${\bf x}$. We proposed RPL as an alternative to softmax, which is based on the open-world assumption. We showed that RPL has beneficial properties for handling the uncertainty concerning the novelty of the input. For application where such a feature is useful, RPL could be an alternative to softmax. We demonstrated that common deep neural network architectures can be trained with RPLs without many modifications. Further research is necessary to understand all the implications of the usage of RPL in depth.

RPL can be used in Bayesian neural networks to combine the desirable properties, e.g., for handling uncertainty and preventing overfitting. We also showed that RPL is less prone to adversarial examples. This can be explained with the open-world assumption inherent in RPL\citep{DBLP:conf/bmvc/RozsaGB17}.

\subsubsection*{Acknowledgments}
% %TODO: uncomment 

The authors acknowledge the financial support by the Federal Ministry of Education and Research of Germany (BMBF) in the project deep.Health (project number 13FH770IX6).

%Use unnumbered third level headings for the acknowledgments. All acknowledgments
%go at the end of the paper. Do not include acknowledgments in the anonymized
%submission, only in the final paper.

%\section*{References}
%References follow the acknowledgments. Use unnumbered first-level heading for
%the references. Any choice of citation style is acceptable as long as you are
%consistent. It is permissible to reduce the font size to \verb+small+ (9 point)
%when listing the references. {\bf Remember that you can use more than eight
%  pages as long as the additional pages contain \emph{only} cited references.}
%\medskip

\small

\bibliographystyle{plainnat}
\bibliography{radial_prediction_layers}

\end{document}

\maketitle

\section{Bayesian Neural Networks}

The Bayesian neural network that we used in this work are trained with a variational approximation and the mean-field approximation~\citep{Blundell15}. The posterior distribution is approximated by a parametrized variational distribution $q({\bf w}\mid \theta)$. $\theta$ is the set of variational parameters.  In the mean field approximation the variational distribution $q({\bf w}\mid \theta)$ factorizes, i.e. $q({\bf w}\mid \theta) = \prod_k q(w_k\mid \theta)$. Here, each $w_k$ (the individual weights respective biases) is represented by a Gaussian distribution $q(w_k\mid \theta)=q(w_k\mid \theta_k)=\mathcal N \left(\mu_k, \sigma_k^2\right)$. So, the variational distribution for  each weight/bias is described by two variational parameters $\theta_k = \{\mu_k, \sigma_k^2\}$. 

In the learning process, the variational parameters for all weights/biases (all $k$) are learned by minimizing the evidence lower bound (ELBO). The learned variational parameter are given by $\theta^* = \text{arg}\min_\theta ELBO(\mathcal D=\{ {\bf X}, {\bf y}\}, \theta)$ with

\begin{equation}\label{elbo-VAE}
  ELBO(\mathcal D=\{ {\bf X}, {\bf y}\}, \theta) =
  D_{KL}[q({\bf w} \mid \theta) \mid \mid p({\bf w})] -
   \sum_i \mathbb E_{q({\bf w}\mid {\bf \theta})}[\log p(y^{(i)} \mid {\bf x}^{(i)}; {\bf w})]
\end{equation}

$D_{KL}$ is the Kullback-Leibler divergence. The sum is over all training examples. The first term on the right hand side forces the approximated posterior  $q({\bf w} \mid \theta)$ to be similar to the prior $p({\bf w})$ (\textit{complexity cost}). The second term depends on the data. It can be interpreted as the term which describes how well the model (depending on ${\bf w}$) predicts the training labels $y^{(i)}$ for the corresponding inputs ${\bf x}^{(i)}$.
%(\textit{likelihood cost}: the expected log-likelihood under the variational approximation).

\subsection{Prediction with Bayesian neural networks}

The general formula for prediction with a Bayesian neural network is:

\begin{equation}
p(y \mid X) = \int p(y \mid X; {\bf w}) \; p({\bf w}\mid \mathcal D_{train}) d{\bf w}
\end{equation}

$p({\bf w}\mid \mathcal D_{train})$ is the posterior of the weights/biases (learned from the training data).
In the variational approximation: $p({\bf w}\mid \mathcal D_{train}) \approx q({\bf w} \mid \theta^*)$. The integral for the prediction is approximated by a Monte-Carlo integration

\begin{equation}
p(y \mid X) = \sum_{l=1}^n p(y \mid X; {\bf w}^{(l)})
\end{equation}

with $n$ samples of the weights/biases ${\bf w} \sim q({\bf w}\mid \theta^*)$. Each of such a complete weights/biases sample, that we also call a sample of the Bayesian neural network. Note, that the individual weights/biases $w_k$ can be drawn independently in the mean-field approximation.

\subsection{Hyperparameter}\label{chp:bayes_hyper}

Typically, a training (update) step is not done with the full dataset but on a small subset (mini-batch). So, the gradient of equation\ref{elbo-VAE} is calculated only with a part of the complete sum. In this case, the first term (\textit{complexity cost}) must be reweighted accordingly. In practice, different weighting schemes between the \textit{complexity cost} and \textit{likelihood cost} are possible. So, the weighting between the terms can be considered as a hyperparameter. We used the hyperparameter $M$ in the following way:

\begin{equation}\label{elbo-VAE-mini-batch}
  \mathcal F(\mathcal D=\{ {\bf X}, {\bf y}\}, \theta) =
  \frac{1}{M} D_{KL}[q({\bf w} \mid \theta) \mid \mid p({\bf w})] -
  \sum_i^{m} \mathbb E_{q({\bf w}\mid {\bf \theta})}[\log p(y^{(i)} \mid {\bf x}^{(i)}; {\bf w})]
\end{equation}
with the KL-weighing hyperparameter $M$ and the minibatch size $m$. As prior we used the scale mixture prior $p({\bf w}) = \prod_k p(w_k)$ with $w_k$ (see~\cite[chapter 3.3]{Blundell15}).

$$
p(w_k) = \pi \mathcal N\left( w_k \mid 0, \sigma_1^2 \right) + (1-\pi) \mathcal N\left( w_k \mid 0, \sigma_2^2 \right)
$$

Hyperparameter values for Bayesian networks summarized:

\begin{itemize}
\item Scale mixture prior with $\pi= 0.35; \sigma_1^2 = 1.0; \sigma^2_2 = 0.0183$
\item Minibatch size $m=50$
\item KL-Weighting Hyperparameter $M=40$
\end{itemize}

\subsection{Bayesian neural networks with Softmax}
\begin{figure}[hbt!]
  %centering
  \includegraphics[width=0.9\linewidth]{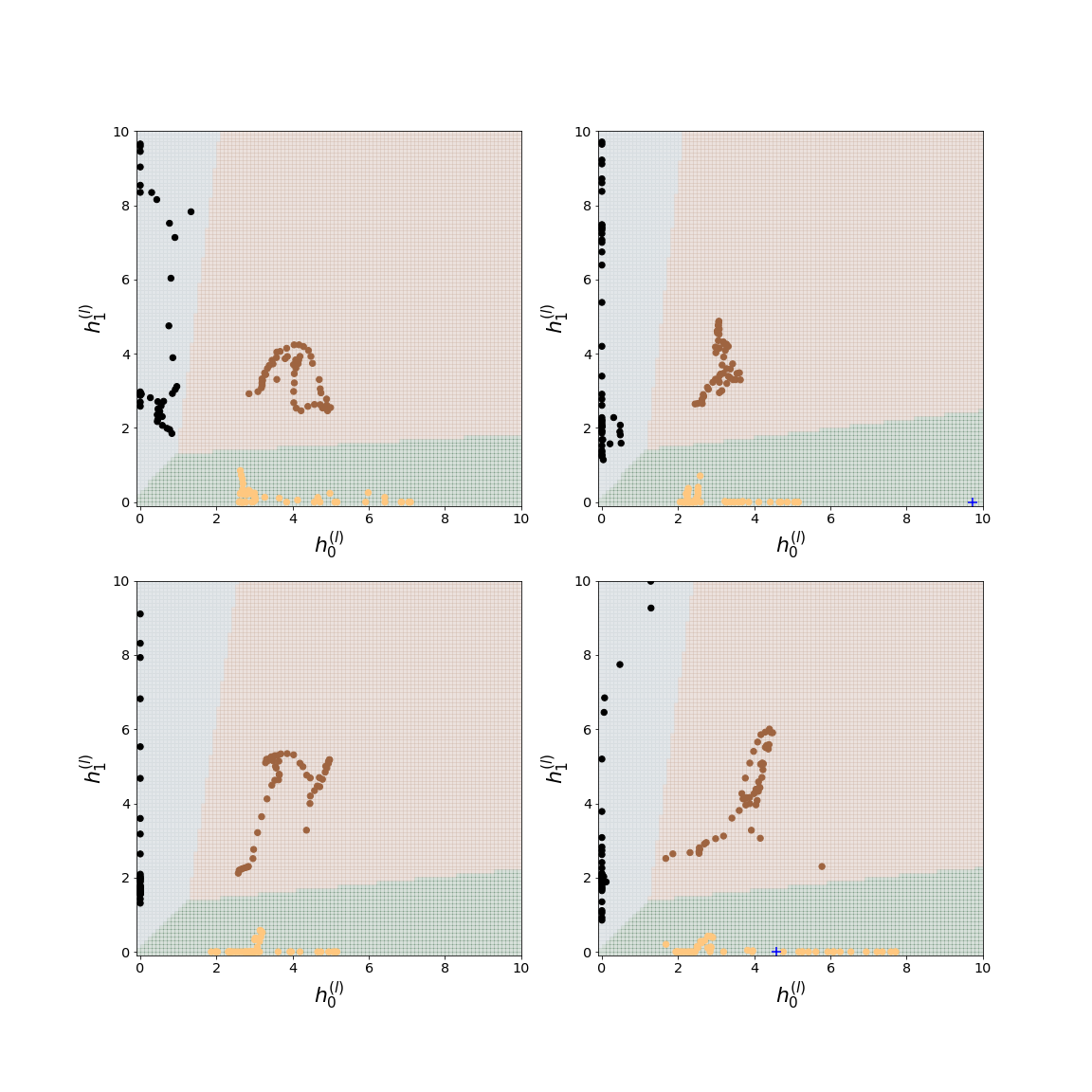}
   \caption{Decision boundaries and mapping of the training data into the last hidden space for the spiral data set.
   The decision boundaries are quite the same for different samples of the Bayesian neural network.}
  \label{fig:spiral_data_last_hidden_softmax_bayes}
\end{figure}

In figure~\ref{fig:spiral_data_last_hidden_softmax_bayes} the mapping into the last hidden space for a network with \textit{softmax} on the sprial dataset is shown. 

\subsection{Bayesian neural networks with RPL}
\begin{figure}[hbt!]
  %centering
  \includegraphics[width=0.9\linewidth]{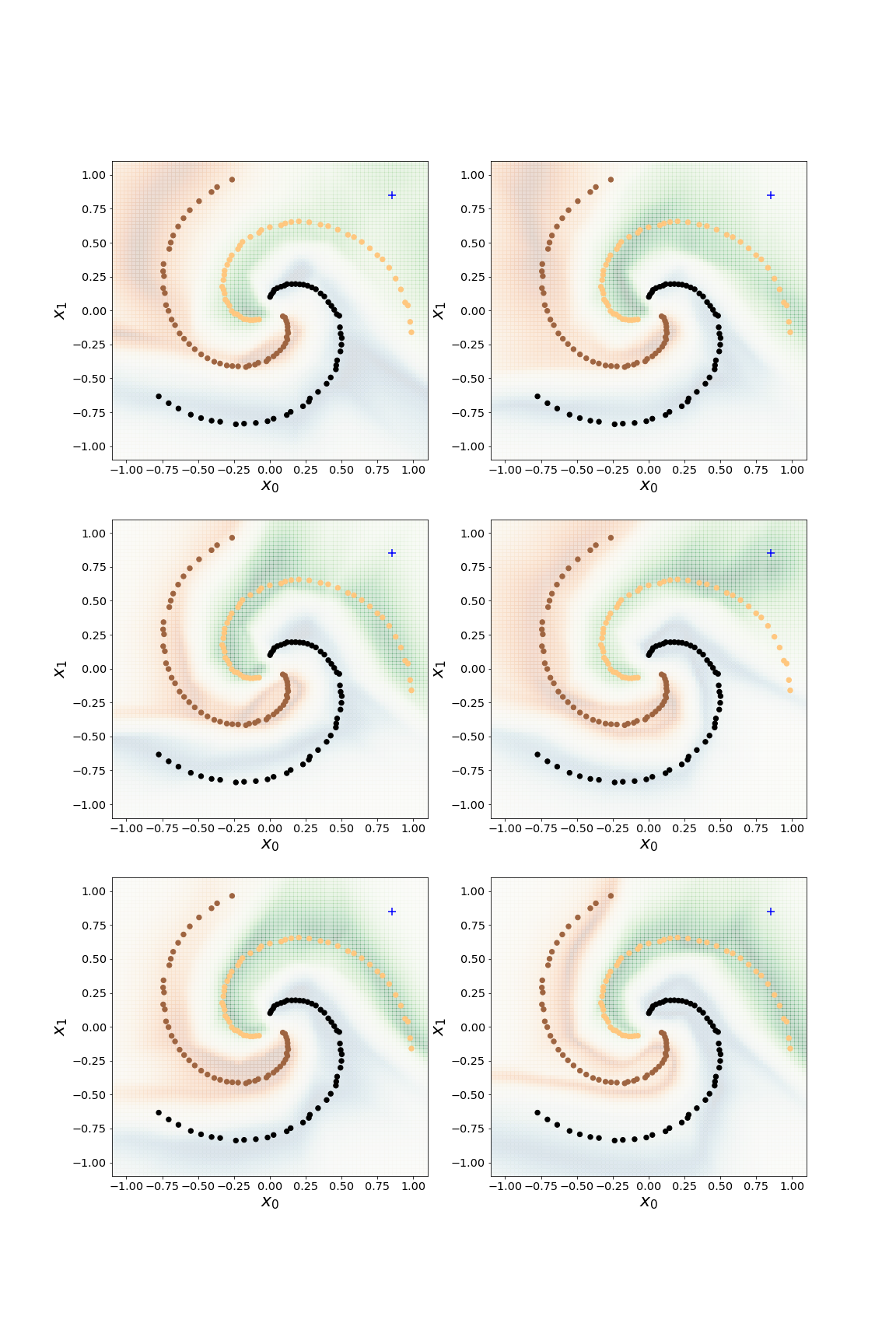}
   \caption{Predictions of six samples from a Bayesian neural network with RPL. The individual samples from the Bayes net
   don't always predict the target classes of the training data with high probabiity. However,
   averaged over the samples }
  \label{fig:rpl_decisions_bayes}
\end{figure}

In figure~\ref{fig:rpl_decisions_bayes} the predictions of six samples of a Bayesian network are shown. 

\section{Training Setups}

\subsection{Spiral and MNIST Data}
Table~\ref{tab:spiral_mnist_architectures} summarizes the network architectures used for the spiral and MNIST datasets. 

For the spiral data we used an RMS-prop optimzier ($\alpha:0.0005$, $\beta:0.9$) and trained the network over 25000 epochs with a mini-batch size of 50. Additional parameters for the Bayesian network are described above~\ref{chp:bayes_hyper}.

To train the MNIST dataset we used a small convolutional neural network with \textit{softmax} and RPL as classifier. Besides we implemented a variation with dropout~\citep{JMLR:v15:srivastava14a} ($p=0.5$) to observe its effect with RPLs. Adam~\citep{kingma2014adam} ($\beta_{1}=0.9$, $\beta_{2}=0.999$, $initial\_learning\_rate = 0.0001$) was used as an optimizer. After some experiments with the RPL hyperparameter $a$ we fixed it to $a=1$, because no effects in terms of the accuracy were observable. But we evaluated several values of the parameter $beta$ to shift the resulting distributions. Every network was trained for 10 epochs with a mini-batch size of $1024$.  

\begin{table}[hbt!]
  \caption{A simple dense feed-forward network was used for the spiral dataset and a small 2 layer convolutional network for the MNIST data.}
  \label{tab:spiral_mnist_architectures}
  \centering
  \begin{tabular}{cccccc}
    \toprule

    \multicolumn{2}{c}{Spiral Data} & \multicolumn{3}{c}{MNIST}                            \\
    \cmidrule(r){1-5}
    Softmax       & RPL          & Softmax        & RPL            & RPL                   \\
    \midrule
    FC-50 + ReLU  & FC-50 + ReLU & Conv5-10       & Conv5-10       & Conv5-10              \\
    FC-50 + ReLU  & FC-50 + ReLU & MaxPool + ReLU & MaxPool + ReLU & MaxPool + ReLU        \\
    FC-2  + ReLU  & FC-50 + ReLU & Conv5-20       & Conv5-20       & Conv5-20 + Drop       \\
    FC-3          & FC-3         & MaxPool + ReLU & MaxPool + ReLU & MaxPool + ReLU        \\
                  &              & FC-100 + ReLU  & FC-100 + ReLU  & FC-100 + ReLU + Drop  \\
                  &              & FC-100 + ReLU  & FC-100 + ReLU  & FC-100 + ReLU         \\
                  &              & FC-10          & FC-10          & FC-10                 \\
    Softmax       & RPL          & Softmax        & RPL            & RPL                   \\
    \bottomrule
  \end{tabular}
\end{table}

\subsection{ILSVRC Data}
We explored RPLs on the ILSVRC dataset (ImageNet)~\citep{Deng09} to verify that the approach is applicable to real-world data and deep neural network architectures. We focused on the VGG network~\citep{Simonyan2014VGG}, which we have studied most intent. The modifications to the VGG network have been kept to a minimum, ensuring comparability to the original results. The softmax layer was, replaced with an RPL. In addition, we removed dropout~\citep{JMLR:v15:srivastava14a} and the reduction of neurons from the fully-connected layers. See Table~\ref{tab:imagenet_architectures} for the full network architecture.

\begin{table}[hbt!]
  \caption{Network architecture VGG.}
  \label{tab:imagenet_architectures}
  \centering
  \begin{tabular}{cc}
    \toprule

    VGG 19\\
    \midrule
    Conv3-64 + BN + ReLU  \\
    Conv3-64 + BN + ReLU  \\
    MaxPool  \\
    Conv3-128 + BN + ReLU  \\
    Conv3-128 + BN + ReLU  \\
    MaxPool  \\
    Conv3-256 + BN + ReLU  \\
    Conv3-256 + BN + ReLU  \\
    Conv3-256 + BN + ReLU  \\
    Conv3-256 + BN + ReLU  \\
    MaxPool  \\
    Conv3-512 + BN + ReLU  \\
    Conv3-512 + BN + ReLU  \\
    Conv3-512 + BN + ReLU  \\
    Conv3-512 + BN + ReLU  \\
    MaxPool  \\
    Conv3-512 + BN + ReLU  \\
    Conv3-512 + BN + ReLU  \\
    Conv3-512 + BN + ReLU  \\
    Conv3-512 + BN + ReLU  \\
    MaxPool + AvgPool \\
    \midrule[0.3pt]
    FC-8192  + ReLU  \\
    FC-8192  + ReLU  \\
    FC-1000   \\
    \midrule[0.3pt]
    RPL   \\
    \bottomrule
  \end{tabular}
\end{table}

The network was trained in two variants: \textit{pre-trained} (reinitialized weights of all fully connected layers) and \textit{from scratch} (reinitialized weigths for all parameters).

Since we were interested in the behavior of RPL in the training process and not in achieving the best possible accuracy value on the ILSVRC dataset we trained with one fixed setup, some exceptions made towards the RPL hyperparameters $a$ and $beta$. We changed the value of $a$ to investigate different lengths ($1$,$2$,$5$,$50$,$100$) of the prototype vectors ${\bf p}_j$ and its effect towards the learning process. No hyperparameter tuning was applied. Each variation was trained for 10 epochs using an Adam optimizer~\citep{kingma2014adam} ($\beta_{1}=0.9$, $\beta_{2}=0.999$, $initial\_learning\_rate = 0.0001$) and a mini-batch size between $128$ (\textit{from scratch)} and $512$ (\textit{pre-trained}).

We used the full ILSVRC dataset (2012 version) to prove deep neural networks using RPL can be trained. We also created two subsets based on ILSVRC dataset to investigate the open world properties of RPLs. On of the subset contains 100 randomly selected classes as training and validation data. The remaining 900 classes were used as novel data in the test phase. For the second subset we removed all classes in the \textit{artifact} category (WorldnetID: 'n00021939') resulting in 478 classes for training and validation phase and 522 classes as novel data points. In terms of the number of samples in this subset, the training and novel data are nearly equal. All data was normalized per color channel using $\mu = (0.485, 0.456, 0.406)$ and $\sigma=(0.229, 0.224, 0225)$. Scripts to create both subsets are provided at: \url{https://gitlab.com/peroyose/radial_prediction_layers}

For the experiments, we used a Server with 4 GPUs (NVIDIA Tesla V100). The model calculation was distributed over the GPUs, splitting the mini-batch into smaller mini-batches and calculate them in parallel. On average it took 6 (\textit{pre-trained}) to 20 (\textit{from scratch}) hours to train a VGG network on the full dataset.

\bibliographystyle{plainnat}
\bibliography{radial_prediction_layers_supplementary}